\documentclass[journal]{IEEEtran}
\ifCLASSOPTIONcompsoc
  \usepackage[nocompress]{cite}
\else
  \usepackage{cite}
\fi
\usepackage{amsmath,amssymb,amsfonts}
\ifCLASSINFOpdf
   \usepackage[pdftex]{graphicx}
   \graphicspath{{../pdf/}{../jpeg/}}
   \DeclareGraphicsExtensions{.pdf,.jpeg,.png}
\else
   \usepackage[dvips]{graphicx}
   \graphicspath{{../eps/}}
   \DeclareGraphicsExtensions{.eps}
\fi
\usepackage{acronym}
\usepackage{multirow}
\usepackage{lscape}
\usepackage{longtable}
\usepackage{gensymb}
\usepackage{booktabs}
\usepackage[table,xcdraw]{xcolor}

\usepackage{supertabular,booktabs}
\usepackage[noend]{algpseudocode}

\usepackage{algorithmicx,algorithm}

\usepackage{array}
\usepackage{mdwmath}
\usepackage{mdwtab}
\usepackage{eqparbox}
\ifCLASSOPTIONcompsoc
  \usepackage[caption=false,font=footnotesize,labelfont=sf,textfont=sf]{subfig}
\else
  \usepackage[caption=false,font=footnotesize]{subfig}
\fi
\usepackage{fixltx2e}
\usepackage{stfloats}
\usepackage{url}
\ifCLASSINFOpdf
  \usepackage[pdftex]{thumbpdf}
\else
  \usepackage[dvips]{thumbpdf}
\fi
\begin{document}
\title{Salient Bundle Adjustment for Visual SLAM}
\author{Ke~Wang,~\IEEEmembership{Member,~IEEE,} 
            ~Sai~Ma,% <-this % stops a space
            ~Junlan~Chen,            
            ~Jianbo~Lu,~\IEEEmembership{Fellow,~IEEE} 
\thanks{K. Wang and S. Ma are with the School of Automobile Engineering, Chongqing University, China, 400044, also with the Key Lab of Mechanical Transmission, Chongqing University, China, 400044 (e-mail: kw@cqu.edu.cn, masai@cqu.edu.cn).}% <-this % stops a space
\thanks{J. Chen is with shool of Economics and Management, Chongqing Normal University, Chongqing 401331, China (e-mail: nwpujunlan@163.com).}% <-this % stops a space
\thanks{J. Lu is with Research and Advanced Engineering, Ford Motor Company, Dearborn, MI 48121 USA (e-mail: jlu10@ford.com).}% <-this % stops a space
\thanks{Manuscript received April 19, 2005; revised August 26, 2015. (Corresponding author: Ke wang)}}


\markboth{Journal of \LaTeX\ Class Files,~Vol.~14, No.~8, August~2015}%
{Shell \MakeLowercase{\textit{et al.}}: Bare Demo of IEEEtran.cls for IEEE Journals}



\maketitle

\begin{abstract}
Recently, the philosophy of visual saliency and attention has started to gain popularity in the robotics community. Therefore, this paper aims to mimic this mechanism in SLAM framework by using saliency prediction model. Comparing with traditional SLAM that treated all feature points as equal important in optimization process, we think that the salient feature points should play more important role in optimization process. Therefore, we proposed a saliency model to predict the saliency map, which can capture both scene semantic and geometric information. Then, we proposed Salient Bundle Adjustment by using the value of saliency map as the weight of the feature points in traditional Bundle Adjustment approach. Exhaustive experiments conducted with the state-of-the-art algorithm in KITTI and EuRoc datasets show that our proposed algorithm outperforms existing algorithms in both indoor and outdoor environments. Finally, we will make our saliency dataset and relevant source code open-source for enabling future research. %https://github.com/Saixiaoma/SBA-SLAM
\end{abstract}


% \fi
\IEEEpeerreviewmaketitle

\section{Introduction}
\IEEEPARstart{S}{LAM}  means to estimate the pose of the robot in unknown environments, simultaneously construct a model of the environment that the sensors are perceiving\cite{RN104, RN783}. SLAM has been investigated in the computer vision and robotics communities over the past few decades, and is still an indispensable module for various applications, ranging from autonomous vehicles\cite{RN185, RN435} and medical robots\cite{RN443, RN371} to augmented and virtual reality\cite{RN441, RN533}. The never-ending quest to come up with real-time, accurate and robust solutions has led to many methods sprung up\cite{RN953, RN881, RN637, RN636, RN940}. 

In general, these methods can be divided into two types: the geometry-based method and the deep learning-based method. Over the past decades, we have seen impressive progress on the geometry-based method and have demonstrated superior performance on accurate and real-time\cite{RN635, RN632, RN598}. However, one of the most important absent features of these methods is the capability to automatically learn knowledge from large-scale dataset. Therefore, these methods are difficult to benefit from the large-scale data available. Recently, deep learning shows the strong capability to handle the redundant information from high-dimensional data. There have been many attempts to utilize deep learning techniques for SLAM/VO\cite{RN685, RN697, RN375, RN702}. Therefore, this paper uses both geometry and deep learning-based techniques in SLAM, since the good complementarity between them. 

Recently, we have witnessed the ongoing evolution of visual SLAM approaches from geometric model-based approaches to deep learning-based approaches\cite{RN58}. Deep learning-based methods have successfully demonstrated the good capability to address the challenging issues such as dynamic objects, illumination changing. However, the progress of these algorithms only relies on the success of deep learning itself, which not explore the internal mechanism relationship between deep learning and SLAM. Some attempts\cite{RN731, RN727, RN425, RN974} have been made to use semantic information to improve the perception of SLAM, such as using semantic prior information to filter dynamic objects. One problem that is difficult to solve is that the semantic information cannot completely determine whether an object is dynamic or static, just like a car can be moving or static. These algorithms treat all selected feature points as equally important in localization and mapping tasks. 

Human performs localization and mapping tasks very differently, which focuses on the most salient objects or features and gives different weight for different features or objects\cite{RN961}. This is called active approach, which actively looks for the most salient features and objects in the field of vision that meet the needs of the task according to the specific task\cite{RN827}. For example, when people walk around in a new environment, they usually focus on the salient landmarks to localize the position. Therefore, very recently, the philosophy of saliency and attention has started to gain popularity in the robotics community\cite{RN778, RN781, RN1020}. For example, Salient-DSO\cite{RN781} uses a saliency model to mimic the qualitative human vision, which selects feature points in salient regions. The saliency model used in this framework that trained on CVL-UMD dataset by using SalGAN\cite{RN948}. However, this saliency model does not completely describe everything the SLAM system should attend to, which makes it unable to work in outdoor environments. This is mainly due to the single focus and center bias of the human gaze dataset. 

Therefore, we first proposed an open-source salient dataset, Salient-KITTI, based on KITTI\cite{RN224}. Different from the human gaze only dataset only focus on human fixations, our proposed dataset also considers the geometric and semantic information. Then, based on Salient-KITTI dataset, we use DI-Net\cite{RN951} to obtain saliency model to predict the region that SLAM system should attend to. Finally, comparing with traditional Bundle Adjustments (BA) method treating all feature as equally important in optimization process, we think that the salient feature points should play more important roles, just like the way humans processing visual information. Therefore, we proposed a Salient Bundle Adjustment (SBA) method to mimic this process. {\bf In summary, our main contributions are as follows:}

\begin{itemize}
\item We proposed a Salient SLAM framework for both indoor and outdoor environments, which can apply in various applications. 
\item We proposed an approach to generate Salient Dataset, and we will also make our Salient-KITTI dataset open-source to facilitate research.
\item We proposed a Salient Bundle Adjustment (SBA) method to mimic human vision system.
\item We provide experimental results on various environments to demonstrate the improved performance of the proposed approach with comparison to the state of the art. 
\end{itemize}

The rest of this paper is organized as follows: In Section \ref{section2}, we review the related work on saliency prediction and SLAM. The Methodology is described in Section \ref{section3}, including saliency prediction and SBA. Experimental results are given in Section \ref{section4}, followed by conclusion in Section \ref{section5}. 
%----------------------------------------------------------------------------------------------------------------------
\section{Related Work}
\label{section2}
\begin{figure*}[!t]
\centering
\includegraphics[width=7in]{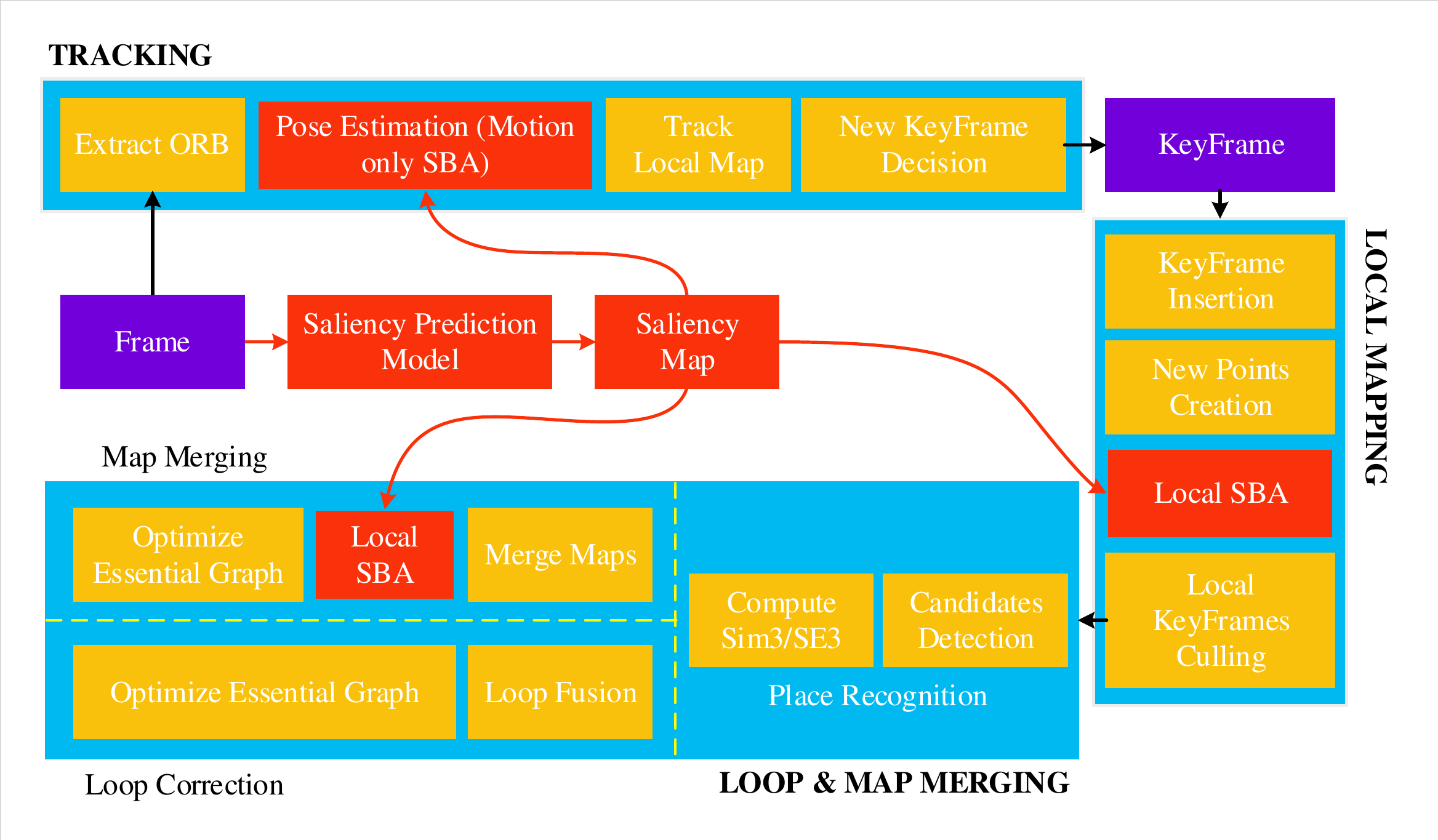}
\DeclareGraphicsExtensions.
\caption{Algorithmic overview of our framework. Red parts: contributions. }
\label{figure1}
\end{figure*}
\subsection{Saliency Prediction}
Bring human visual saliency mechanism to computer vision, it can not only allocate computing resources to important objects but also can produce results that more like the human visual perception. Therefore, visual saliency prediction has become a research hotspot and has received great attention\cite{RN827, RN952, RN1061, RN961}. According to human visual and attention mechanism, the saliency prediction algorithm can be divided as bottom-up and top-down algorithms. One of the earliest works in saliency prediction was proposed by Itti et al.\cite{RN962}. They proposed a visual saliency mechanism-based model by utilizing the feature integration theory and guided search model. After then, the computer vision community has proposed large number of visual saliency calculation models, which predict the possibility of the human eye staying at a certain position in the scene\cite{RN1057, RN835, RN1054}. These classical algorithms are mostly concerned with the bottom-up visual saliency mechanism, which utilizes the color, brightness, and edge features to predict a saliency map. 

Recently, deep learning-based saliency prediction model have achieved great improvements\cite{RN946, RN942, RN1068, RN1066}. The early attempt to applied deep learning in saliency prediction is Vig et al.\cite{RN945}. They use the feature learned from neural network and SVM to classify the patch to be salient or non-salient. Since then, many effective and complex models were proposed, and achieved great improvement in terms of accuracy. For example, DeepNet\cite{RN949}, SALICON-Net\cite{RN999}, SalGAN\cite{RN948}, DeepVS\cite{RN946}, ACL-Net\cite{RN961}, and DeepFix\cite{RN942} explore different network architectures and learning techniques to learn more representative features to make the system closer to human. 

In field of autonomous vehicle, there has been some focus on driver saliency prediction to parse the driver’s attention behaviors as well as road scene to predict the potential unsafe maneuvers\cite{RN778, RN1020}. Deng et al.\cite{RN778} collected an eye tracking dataset in driving scenes and proposed a convolutional-deconvolutional neural network to predict the saliency in driving scenes. Pal et al.\cite{RN1020} proposed a Semantic Augmented Gaze detection approach to predict the saliency regions by considering the depth, vehicle speed, and pedestrian crossing intent. However, these models are more focused on vehicle, pedestrian, and road, etc. The feature point in these areas are not stable and robust in SLAM/VO tasks due to the world static assumption. We hope the saliency model more focus on texture rich and stable areas such as road marking line, traffic lights, traffic signs, and ignore dynamic objects. Therefore, we proposed a saliency prediction model incorporating geometric and semantic information to solve this problem. 
\subsection{SLAM}
\begin{figure*}[htbp]
\centering
\includegraphics[width=7in]{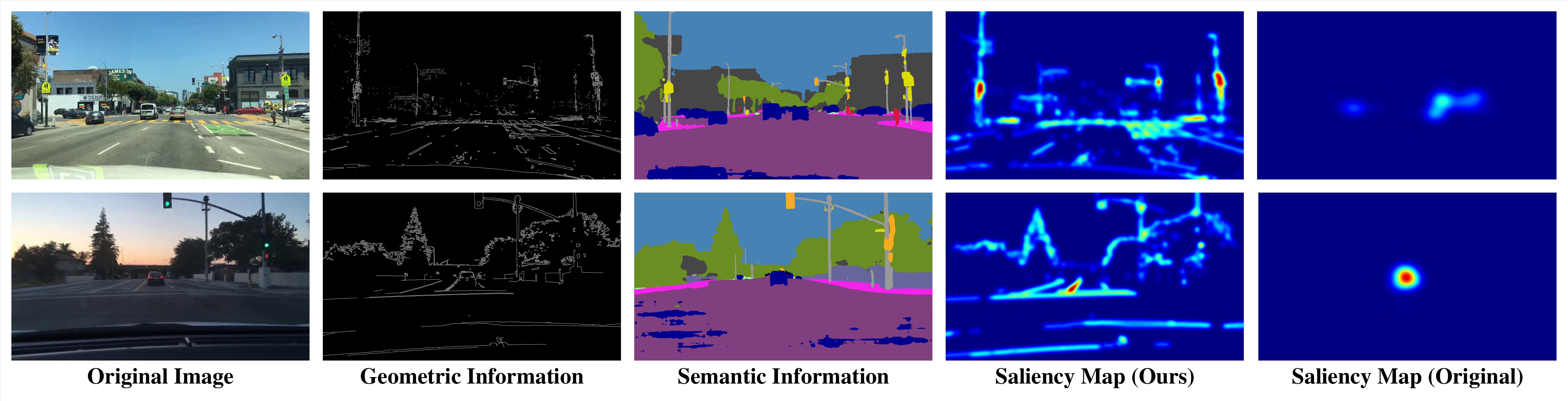}
\DeclareGraphicsExtensions.
\caption{Comparison of our proposed semantic gaze with human gaze-only ground-truth. }
\label{figure2}
\end{figure*}

Generally, geometry-based method can be divided into two categories: feature based method, and direct method. Feature based methods estimate camera pose by extracting and matching interest points from image while direct methods directly use pixel intensity in image to estimate camera pose by minimize the photometric errors. MonoSLAM\cite{RN630} was the first real-time visual SLAM system by using an Extended Kalman Filter (EKF) and Shi-Tomasi points. PTAM\cite{RN631} used the nonlinear optimization method to instead filter based method as back-end optimization methods, and it also proposed and realized the parallelization of tracking and mapping process. ORB-SLAM\cite{RN632} use more stable and effective ORB feature and use three threads to build a complete SLAM: tracking, mapping, and loop detection thread. This system still is one of the most successfully SLAM system by now, and has been extended in \cite{RN953, RN598}. Since cannot extract enough feature points in texture-less environments, many algorithms are also attempt to use line and plane features\cite{RN623, RN604, RN621}. 

Direct method does not extract feature points, but use pixel intensity to estimate camera pose by minimizing the photometric errors. DTAM\cite{RN634} was the first system using direct method to generate a dense 3D map. However, it needs commercial GPU to perform computing. To improve efficient, SVO\cite{RN635} extract FAST features to estimate camera pose and 3D structure by using direct manner. Then, LSD-SLAM\cite{RN636} extended this work, and it can generate semi-dense maps in large-scale environments. DSO\cite{RN637} achieve a trade-off between accuracy and efficient by jointly optimizing all involved model parameters, including geometry parameter, camera pose and intrinsics. Extensions of this work include stereo\cite{RN636} and loop-closing\cite{RN1043}. 

Recently, deep learning based SLAM/VO have achieved great improvements, relying on the powerful learning ability of neural network\cite{RN1040}. PoseNet\cite{RN354} was one of the earliest works that uses CNN to estimate camera pose in an end-to-end manner. After that, many studies have sprung up\cite{RN881, RN955, RN958, RN375}. Deep VO\cite{RN403} use RNN to model the relationship of motion dynamic and image sequences. ESP-VO\cite{RN375} then extended this work by adding uncertainty estimation for pose estimation. To avoid collecting tremendous data, many studies tried to use unsupervised techniques\cite{RN348, RN357, RN387}. \cite{RN200} proposed a temporal (forward-backward) and spatial (left-right) constraints to learn consistent pose and 3D structures, and better exploit unsupervison. Almalioglu et al.\cite{RN394} use generative adversarial networks (GANs) for camera pose estimation.

Besides, many researchers attempt to combine geometry-based method with deep learning-based method\cite{RN1026, RN425, RN974}. \cite{RN1026} utilizes deep learning-based semantic segmentation and scene parsing techniques to reduce the ambiguity of static and dynamic regions of the image. DynaSLAM\cite{RN1045, RN1044} utilize both deep learning and multi view geometry to segment dynamic objects. DS-SLAM\cite{RN974} combined semantic segmentation and moving consistency check to reduce the influence of dynamic objects to improve the accuracy in dynamic environments. Later, Salient-DSO\cite{RN781} extracted feature points in salient regions by using the mechanism of visual saliency and attention. This method enhanced the performance of DSO. However, this framework can only work in indoor environments. Therefore, we extended this method to outdoor environments by using our proposed saliency model and salient-BA approach.
\section{Methodology}
\label{section3}
\subsection{System overall}
\label{section3.1}

Figure \ref{figure1} provides a simplified illustration of our proposed framework, which comprises of two components: a geometry-based SLAM pipeline, and a learning-based saliency prediction module. The saliency prediction module generates saliency map corresponding to SLAM dataset. Then, the saliency maps are used as input to help SLAM choose the salient feature points to improve the accuracy and robustness. In our proposed framework, we adopt ORB-SLAM3\cite{RN953} as the backbone. In rest of Section \ref{section3}, we begin by firstly describing how to make a salient dataset and use it to obtain saliency prediction model in Section \ref{section3.2}. Next, in Section \ref{section3.3}, we describe how to design the Salient Bundle Adjustment (SBA) by using saliency maps predicted by our proposed saliency model. 

\subsection{Saliency prediction}
\label{section3.2}
\begin{figure*}[htbp]
\centering
\includegraphics[width=7in]{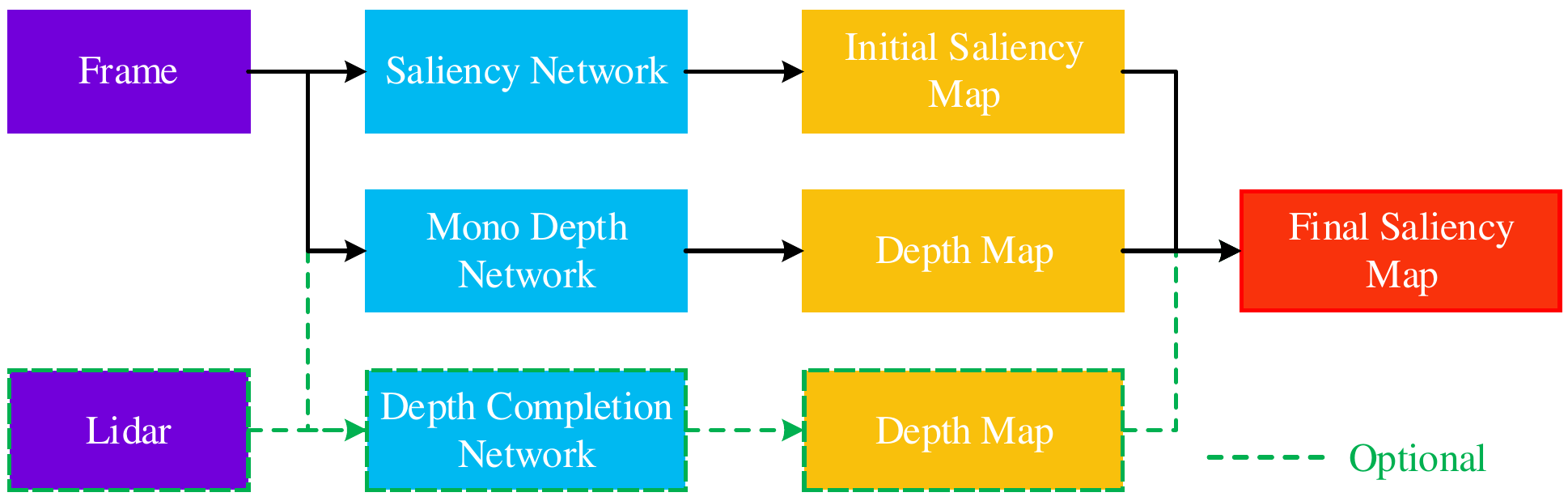}
\DeclareGraphicsExtensions.
\caption{The pipeline of visual saliency map prediction. }
\label{figure3}
\end{figure*}
Visual saliency or visual attention mechanism refers to mimic the human vision system to select the most salient and interested regions or points from natural scenes for further processing under different tasks. This is curial for SLAM/Odometry, where the system can pay more attention to areas with rich corners and lines, while ignoring the dynamic dynamics, in complex environments. Recently, there are many deep learning-based techniques to predict saliency areas in natural scenes, and achieved excellent results. However, these saliency prediction approaches not completely describe everything the VO/SLAM system should pay attention to, which is mainly due to the training dataset have center-bias. The reason is that these approaches only uses raw human gaze information, while human gaze will stay on the road in front of vehicle because this is where the vehicle goes. However, this is not enough, because SLAM/VO also needs to focus on the areas away from the center of the image. Relying only gaze data obtained by human eye tracker does not help capture all these important cues. 

Therefore, to solve this problem, we adopt the similar strategy as \cite{RN1020}, and make a salient dataset to train saliency model by combining both geometric and semantic information, which uses semantic gaze to instead human gaze ground-truth. Our proposed salient dataset is based on KITTI Object Detection Dataset\cite{RN224}. Specifically, we firstly extract geometric information for each image such as feature points, lines and planes. The reason is that the classical SLAM/VO approach usually focus on the area with rich geometric information. Next, we use SDC-Net\cite{RN1046} to generate segmented mask around the interest object. We select 13 categories (traffic light, traffic sign, road, building, sidewalk, parking, rail track, fence, bridge, pole, pole-group, vegetation, terrain) to filter the geometric information, which regions of these categories usually contains salient, stable, and robust features. Therefore, the feature in dynamic regions are not appear in our salient dataset such as moving vehicles, person, rider, etc. Figure \ref{figure2} show the comparison of our proposed semantic gaze and human gaze ground-truth. Finally, based on our proposed salient dataset, we can obtain a saliency model by using DI-Net\cite{RN951},and use it to predict an initial saliency map. 

Besides, we also consider that the distance between the objects and ego-vehicle will impact the saliency. Just as humans pay more attention to objects that are closer to themselves, because they are more likely to interact with objects that are closer. Therefore, we use monocular depth estimation network, MonoDepth\cite{RN1047}, to generate depth map for correct saliency map. Besides, if Lidar or other depth sensors available, we can use depth completion network to obtain more accurate depth map. Figure \ref{figure3} show the pipeline of our saliency prediction. All steps to generate saliency map are summarized in Algorithm \ref{algorithm1}. 

\begin{algorithm}[t]
\caption{The Algorithm for saliency map prediction. }
\label{algorithm1}
\hspace*{0.02in}{\bf Input:}
Input image I, Input Lidar point L (optional)\\
\hspace*{0.02in}{\bf Output:}
Prediction Final Saliency Map $S_{(final)}$
\begin{algorithmic}[1]
\State Extract geometric information for each image.
\State $SM = SDC - Net(I)$; $SM: segmented mask$.
\State Utilize $SM$ to filter geometric information.
\State Make salient dataset and train saliency prediction model $(DI-Net)$.
\State ${S_{(init)}} = DI - Net(I)$
\State $D = MonoDepth(I) or D = DFuseNet(L + I)$; $D:$ depth map.
\For{${\forall ({x_i},{x_j}) \in I}$}
      \State ${\widehat S_{(final)}} = a{S_{(init)}}\left( {{x_i},{x_j}} \right)/D\left( {{x_i},{x_j}} \right) + b$
\EndFor
\State ${{S_{(final)}} = Normalization({\widehat S_{(final)}},0,255)}$
\end{algorithmic}
\end{algorithm}
\subsection{Salient Bundle Adjustment}
\label{section3.3}
 In section \ref{section3.2}, we obtained the saliency map for each frame by using saliency prediction model. In this section, we will describe how saliency maps used in pose estimation process. In classical SLAM or VO system, we usually perform BA (Bundle Adjustment) to optimize the camera ego motion in tracking thread (motion-only BA), and to optimize the local window of keyframes and map-points in local mapping thread (Local BA). If loop closure is available, it also optimizes all keyframes and map-points after a loop closure (full BA). In traditional BA algorithm, all feature points are treated as equally, which make the most salient feature point cannot do more. Therefore, we proposed Salient Bundle Adjustment approach to make the salient feature points full play to its value. The salient BA defined as follows: 
\begin{figure*}[htbp]
\centering
\includegraphics[width=7in]{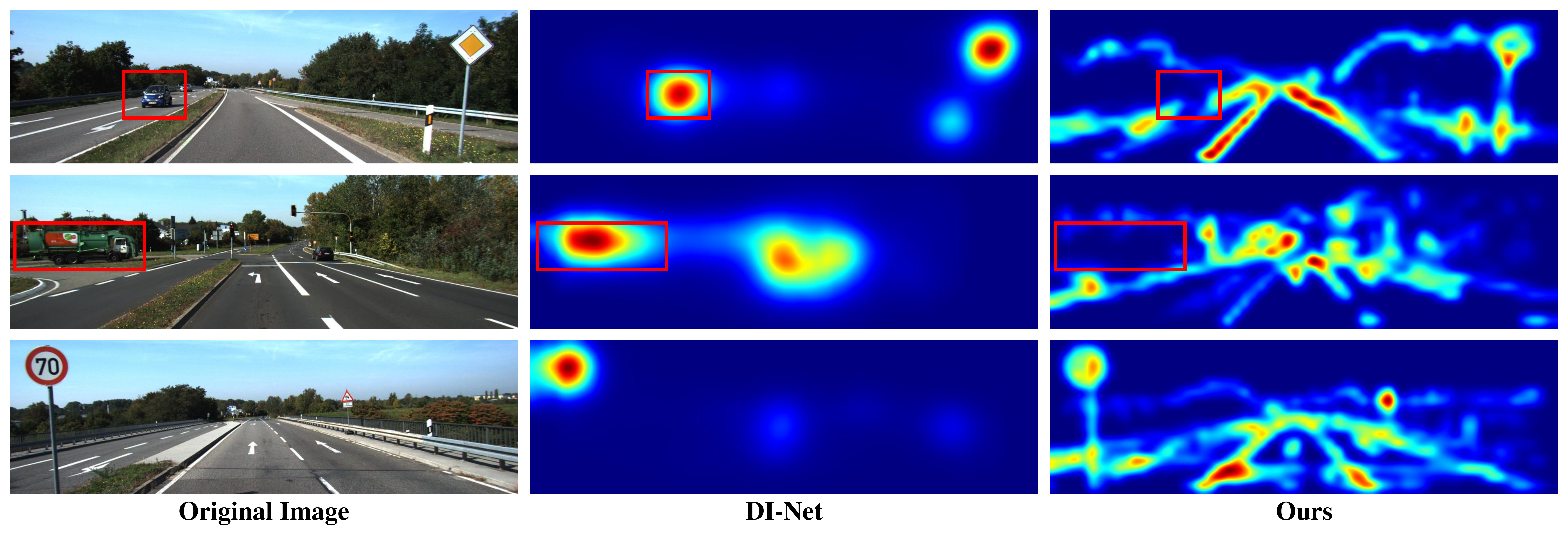}
\DeclareGraphicsExtensions.
\caption{Comparison of saliency prediction of saliency model trained in SALICON and Salient-KITTI dataset.}
\label{figure4}
\end{figure*}

 {\bf Motion-only BA:} refers to optimize the camera orientation ${{\bf R}} \in SO(3)$ and translation ${\bf t} \in {\mathbb{R}^3}$, and minimizing the reprojection errors between matched 3D points ${{\bf X}^{i}} \in {\mathbb{R}^3}$  in world coordinates and key-points ${{\bf x}^{i}_{m}} \in {\mathbb{R}^2}$  (monocular camera) or ${{\bf x}^{i}_{s}} \in {\mathbb{R}^3}$ (stereo camera) in camera pixel coordinates, with $i \in \mathcal{X}$ the set of all matches. 
 $$\{ {\bf R,t}\}  = \mathop {\arg \min }\limits_{{\bf R,t}} \sum\limits_{i \in \mathcal{X}} \rho  \left( {{w_i}\left\| {{\bf x}_{( \cdot )}^i - {\pi _{( \cdot )}}\left( {{\bf R}{{\bf X}^i} + {\bf t}} \right)} \right\|_\Sigma ^2} \right)$$
 
 Where $\rho $  is the robust Huber cost function and  $\sum $ the covariance matrix associated to the scale of the key-point. The reprojection function  ${\pi _{( \cdot )}}$ are defined as follows: 
 $${\pi _{\text{m}}}\left( {\left[ {\begin{array}{*{20}{c}}
   X  \\ 
   Y  \\ 
   Z  \\ 
\end{array} } \right]} \right) = \left[ {\begin{array}{*{20}{c}}
\begin{gathered}
   {{f_x}\frac{X}{Z} + {c_x}} \hfill  \\ 
   {{f_y}\frac{Y}{Z} + {c_y}} \hfill  \\ 
   \end{gathered}
\end{array} } \right]$$
$${\pi _s}\left( {\left[ {\begin{array}{*{20}{c}}
   X  \\ 
   Y  \\ 
   Z  \\ 
\end{array} } \right]} \right) = \left[ {\begin{array}{*{20}{c}}
      \begin{gathered}
   {{f_x}\frac{X}{Z} + {c_x}} \hfill \\ 
  {f_y}\frac{Y}{Z} + {c_y} \hfill \\
  {f_x}\frac{{X - b}}{Z} + {c_x} \hfill \\ 
\end{gathered}   \\ 
\end{array} } \right]$$
 Where $\left( {{f_x},{f_y}} \right)$ is the focal length, $\left( {{c_x},{c_y}} \right)$is the principal point, and $b$ is the baseline. 
 $w_{i}$ is the salient weight, which defined as follow: ${{w_i} = a{S^2}\left( {{x_i},{y_i}} \right) + b}$, $S( \cdot )$ is the pixel value in saliency map, $a, b$ are constant value. 

{\bf Local BA:} refers to optimize a set of co-visible keyframe ${\mathcal{K}_L}$ and all map-points in these keyframes ${\mathcal{P}_L}$. Defining ${\mathcal{X}_k}$ as the set of matches between points in ${\mathcal{P}_L}$ and key-points in a keyframe $k$, the optimization problem can be defined as:
 $${\left\{ {{{\bf X}^i},{{\bf R}_l},{{\bf t}_l}|i \in {\mathcal{P}_L},l \in {\mathcal{K}_L}} \right\} = \mathop {\arg \min }\limits_{{{\bf X}^i},{{\bf R}_l},{{\bf t}_l}} \sum\limits_{k \in {\mathcal{K}_L} \cup {\mathcal{K}_F}} {\sum\limits_{j \in {\mathcal{X}_k}} {\rho \left( {E_{k,j}} \right)} }} $$
 $$E_{k,j} = {w_i}\left\| {{\bf x}_{( \cdot )}^j - {\pi _{( \cdot )}}\left( {{{\bf R}_k}{{\bf X}^j} + {{\bf t}_k}} \right)} \right\|_\Sigma ^2$$

 \section{Evaluation}
 \label{section4}
 In this section, we validate our proposed algorithm in two aspects. Firstly, we compare the algorithm that uses the saliency model trained on SALICON\cite{RN1048} and Salient-KITTI dataset, respectively. Secondly, we evaluate our proposed algorithm on KITTI\cite{RN224} and EuRoc\cite{RN213} dataset, and compare our proposed algorithm with state of the art techniques. During experiments, it runs on a computer with Nvidia GeForce GTX 1650 GPU, AMD Ryzen 5-2600 CPU, and 8 GB memory in Linux. 
 \subsection{Dataset used in evaluation}
 KITTI dataset: contains real-world image data that collected from scenes such as urban, rural and highways areas. Each image can contain up to 15 cars and 30 pedestrians, with various degrees of occlusion and truncation. For odometry/SLAM, this dataset provides 22 sequences under autonomous vehicle scenes, with a total length of 39.2 km, and provide 11 sequences (00 - 10 sequence) with ground truth. 

EuRoc dataset: is a monocular, stereo and IMU dataset for indoor MAV. It contains two scenes, one is an industrial environment and the other is an ordinary room environment. This dataset provides 11 sequences with ground truth. And these sequences are divided as three levels of difficulty according to the texture quality, scene brightness, and the speed of motion. 
\subsection{Saliency model evaluation}
{\linespread{1.2}
\begin{table*}[]
\caption{The comparison of saliency prediction model. }
\label{table1}
\centering
\begin{tabular}{ccccccccccccc}
\hline
\multirow{3}{*}{Seq.} & \multicolumn{6}{c}{RPE (Monocular)}                                           & \multicolumn{6}{c}{RPE (Stereo)}                                              \\ \cline{2-13} 
                      & \multicolumn{2}{c}{RMSE[m]} & \multicolumn{2}{c}{MEAN[m]} & \multicolumn{2}{c}{STD} & \multicolumn{2}{c}{RMSE[m]} & \multicolumn{2}{c}{MEAN[m]} & \multicolumn{2}{c}{STD} \\ \cline{2-13} 
                      & DI-Net      & Ours       & DI-Net      & Ours       & DI-Net     & Ours       & DI-Net      & Ours       & DI-Net      & Ours       & DI-Net     & Ours       \\ \hline
0                     & 0.2345      & {\bf 0.1871}     & 0.1456      & {\bf0.1253}     & 0.1839     &{\bf 0.1389}     & 0.0290       & {\bf0.0275}     & 0.0195      & {\bf0.0190}      & 0.0215     & {\bf0.0201}     \\
1                     & --          & --         & --          & --         & --         & --         & 0.0506      & {\bf0.0503}     & 0.0453      & {\bf0.0438}     & {\bf0.0226}     & 0.0305     \\
2                     & 0.1578      & {\bf0.1522}     & 0.1257      & {\bf0.1237}     & 0.0954     & {\bf0.0937}     & 0.0291      & {\bf0.0284}     & 0.0233      & {\bf0.0232}     & 0.0174     & {\bf0.0159}     \\
3                     & 0.0380       & {\bf0.0374}     & 0.0303      & {\bf0.0287}     & {\bf0.0229}     & 0.0236     & 0.0178      & {\bf0.0172}     & 0.0153      & {\bf0.0148}     & 0.0090      & {\bf0.0086}     \\
4                     & 0.0852      & {\bf0.0693}     & 0.0681      & {\bf0.0597}     & 0.0504     & {\bf0.0352 }    & 0.0212      & {\bf0.0203}     & 0.0189      & {\bf0.0179}     & 0.0096     & {\bf0.0091}     \\
5                     & --          & {\bf0.5288}     & --          & {\bf0.3917}     & --         & {\bf0.3549}     & 0.0166      & {\bf0.0165}     & 0.0129      & {\bf0.0128}     & 0.0105     & {\bf0.0099}     \\
6                     & 0.6626      & {\bf0.6246}     & 0.4490       & {\bf0.4245}     & 0.4871     & {\bf0.4575}     & 0.0207      & {\bf0.0185}     & 0.0146      & {\bf0.0140}      & 0.0146     & {\bf0.0122}     \\
7                     & 0.5754      & {\bf0.5407}     & 0.4117      &{\bf 0.3843}     & 0.4019     & {\bf0.3800}       & 0.0168      & {\bf0.0162}     & 0.0139      & {\bf0.0137}     & 0.0094     &{\bf 0.0089}     \\
8                     & 0.2192      &{\bf 0.2160}      & 0.1591      & {\bf0.1580}      & 0.1506     & {\bf0.1470}      & 0.0392      & {\bf0.0389}     & 0.0256      & {\bf0.0253}     & 0.0297     & {\bf0.0296}     \\
9                     & 0.3551      & {\bf0.2858}     & 0.2150       & {\bf0.1948}     & 0.2820      & {\bf0.2090}      & 0.0218      & {\bf0.0214}     & 0.0187      & {\bf0.0183}     &{\bf 0.0112}     & 0.0113     \\
10                    & 0.0673      &{\bf 0.0665}     & 0.0542      & {\bf 0.0537 }    & 0.0399     & {\bf0.0390}      & 0.0206      & {\bf0.0200}       & 0.0153      & {\bf0.0151 }    & 0.0139     & {\bf0.0134}     \\ \hline
\end{tabular}
\end{table*}
}
To validate the effectiveness of our proposed Salient KITTI dataset, we firstly design an experiment to compare the saliency maps generated by the saliency prediction model trained on SALICON and Salient KITTI dataset, respectively. We provide some representative results in Figure \ref{figure4}. For the model trained on SALICON dataset, when there are no significant objects in the image, the focus of attention in concentrated in the center of the image, thereby ignoring other important information. On the contrary, the model trained on Salient KITTI dataset can successfully capture this important information. Moreover, these models can reduce the impact of dynamic objects, so that the points with high saliency usually are stable and robust points, shown in figure \ref{figure4} red boxes. 
\begin{figure}[htbp]
\centering
\includegraphics[width=3.5in]{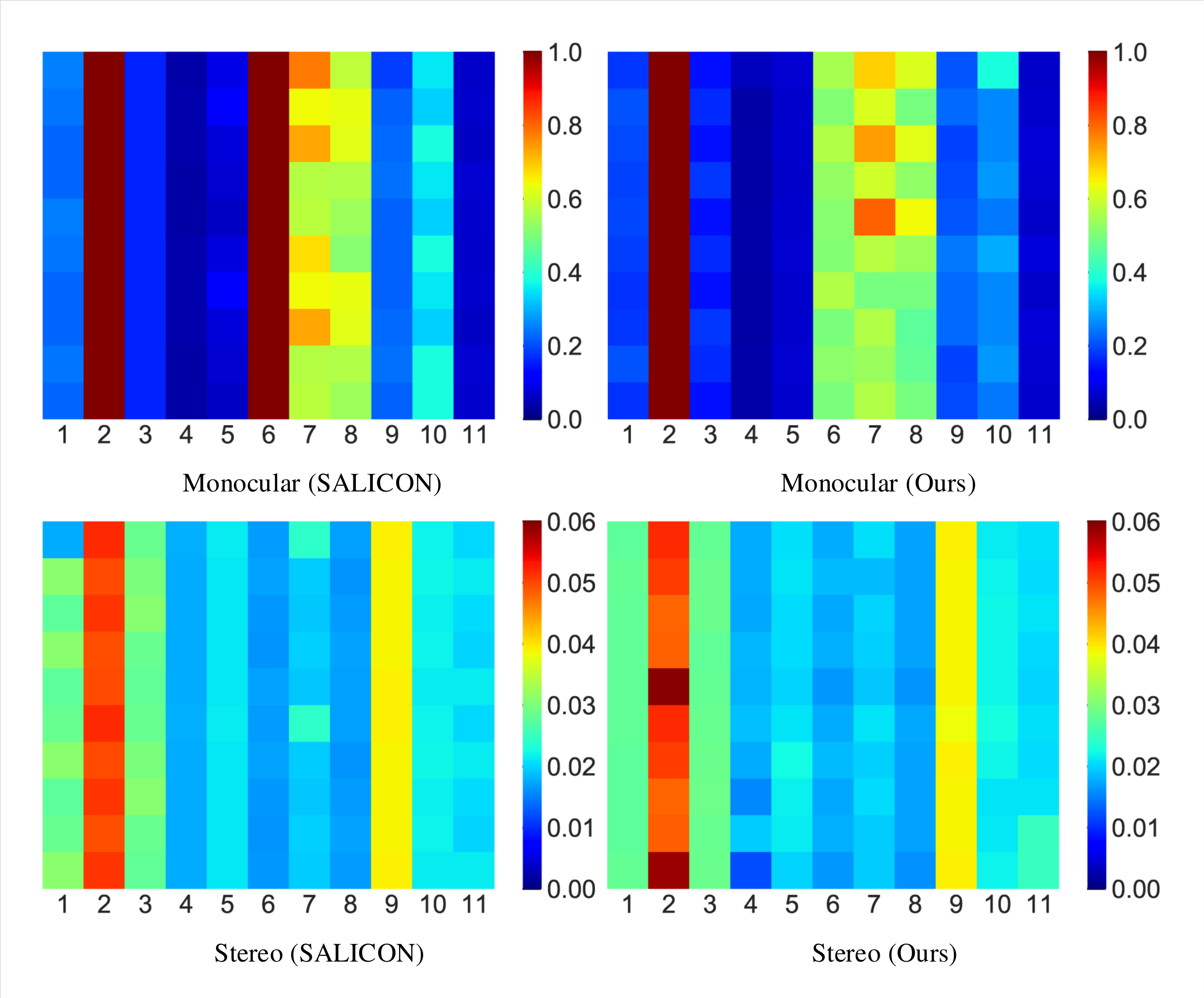}
\DeclareGraphicsExtensions.
\caption{Colored squares represent the RMSE for ten different execution in each sequence of the KITTI dataset. }
\label{figure5}
\end{figure}

We then use these two saliency models in our proposed framework to evaluate its performance. In this experiment, we set two sensors configurations: monocular and stereo camera. We use evo\cite{RN1049} to evaluate algorithms with RPE (relative pose error). The results are listed in Table \ref{table1}, and the correspond heatmaps are shown in Figure \ref{figure5} (each rectangle shows a different run). 

As shown in Table \ref{table1}, we can find the algorithm using saliency model trained on Salient-KITTI dataset achieves more accurate results than the algorithm using the saliency model trained on SALICON dataset, in most cases by a wide margin especially in monocular sensor configuration. In sequence 05, the saliency model trained on SALICON dataset track lost while our proposed saliency model can work well under monocular configuration. The reason is that the most frame has no salient objects leading the attention focused in the center of images. They cannot capture other important information in surrounding areas, and cannot reduce the influence of dynamic objects. 

\subsection{Evaluation on KITTI dataset}
\label{section4.3}
\begin{figure}[htbp]
\centering
\includegraphics[width=3.5in]{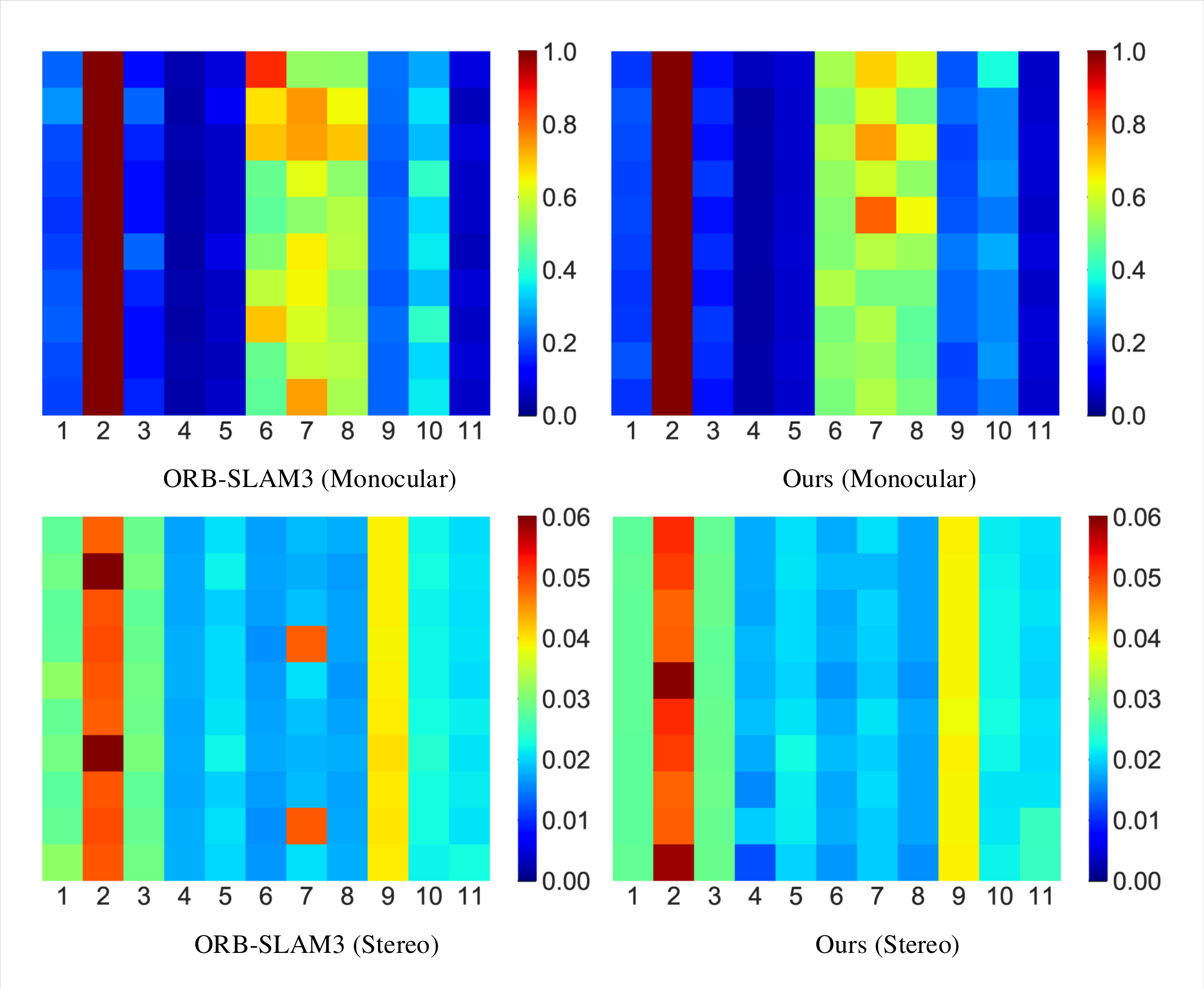}
\DeclareGraphicsExtensions.
\caption{Colored squares represent the RMSE for ten different execution in each sequence of the KITTI dataset.}
\label{figure6}
\end{figure}

  In this section, we design an experiment to validate the effectiveness of SBA in ORB-SLAM3. We also use evo to evaluate algorithms with RPE (relative pose error) in monocular and stereo camera configuration, and to plot trajectory. Experimental results are shown in Table \ref{table2}, which shows the RPE of various sequences. Besides, the full set of results for each evaluated trajectory error is visualized in Figure \ref{figure6}. 

To summarize performance, we have run these algorithms 10 times in each sequence. Because the median value is not affected by the maximum or minimum value of the distribution sequence, the representativeness of the median value to the distribution sequence is improved to a certain extent. Therefore, we chose the median of 10 executions as the finally result. In both monocular and stereo configurations, our system is more accurate than ORB-SLAM3 due to the SBA that makes the salient feature points full play its value. Therefore, this experiment demonstrated that using saliency map can make our algorithm have more advantages in pose estimation. The intuition behind this is that the visual saliency contains high-level semantic information which inherently makes the feature more robust. And our proposed SBA approach further strengthens the role of these saliency points. Besides, our saliency model inference an image only take 8 ms, the whole system can still run in real-time in our platform. 
{\linespread{1.2}
\begin{table*}[htbp]
\caption{The comparison of our proposed algorithm with ORB-SLAM3. }
\label{table2}
\centering
\begin{tabular}{ccccccccccccc}
\hline
\multirow{3}{*}{Seq.} & \multicolumn{6}{c}{RPE (Monocular)}                                                                & \multicolumn{6}{c}{RPE (Stereo)}                                                                   \\ \cline{2-13} 
                      & \multicolumn{2}{c}{RMSE{[}m{]}} & \multicolumn{2}{c}{MEAN{[}m{]}} & \multicolumn{2}{c}{STD{[}m{]}} & \multicolumn{2}{c}{RMSE{[}m{]}} & \multicolumn{2}{c}{MEAN{[}m{]}} & \multicolumn{2}{c}{STD{[}m{]}} \\ \cline{2-13} 
                      & ORB3           & Ours           & ORB3           & Ours           & ORB3           & Ours          & ORB3           & Ours           & ORB3           & Ours           & ORB3           & Ours          \\ \hline
0                     & 0.2087         & {\bf 0.1871}         & 0.1384         & {\bf0.1253}         & 0.1562         & {\bf0.1389}        & 0.0289         & {\bf0.0275 }        & 0.0194         & {\bf0.0190}          & 0.0213         & {\bf0.0201}        \\
1                     & --             & --             & --             & --             & --             & --            & 0.0516         &{\bf 0.0503}         & 0.0450          &{\bf 0.0438}         & {\bf0.0243}         & 0.0305        \\
2                     & 0.1629         & {\bf0.1552}         & {\bf0.1233}         & 0.1237         & 0.1048         & {\bf0.0937}        & 0.0288         & {\bf0.0284}         & 0.0234         &{\bf 0.0232}         & 0.0169         & {\bf0.0159}        \\
3                     & 0.0384         & {\bf0.0374}         & 0.0300           & {\bf0.0287}         & 0.0239         & {\bf0.0236}        & 0.0176         & {\bf0.0172}         & 0.0153         & {\bf0.0148}         & 0.0088         & {\bf0.0086}        \\
4                     & 0.0714         & {\bf0.0693}         & {\bf0.0583}         & 0.0597         & 0.0407         & {\bf0.0352}        & 0.0207         & {\bf0.0203}         & 0.0186         & {\bf0.0179}         & 0.0092         & {\bf0.0091}        \\
5                     & 0.6090          &{\bf 0.5288}         & 0.4229         & {\bf0.3917}         & 0.4319         &{\bf 0.3549}        & 0.0167         & {\bf0.0165}         & 0.0131         & {\bf0.0128}         & 0.0105         & {\bf0.0099}        \\
6                     & 0.6294         & {\bf0.6246}         & 0.4267         & {\bf0.4245}         & 0.4624         & {\bf0.4575}        & 0.0250          & {\bf0.0185}         & 0.0149         &{\bf 0.0140}          & 0.0192         & {\bf0.0122}        \\
7                     & 0.5785         & {\bf0.5407}         & 0.4104         &{\bf 0.3843}         & 0.4075         & {\bf0.3800}          & 0.0172         &{\bf 0.0162}         & 0.0140         & {\bf0.0137}         & 0.0098         & {\bf0.0089}        \\
8                     & 0.2242         & {\bf0.2160}          & 0.1646         & {\bf0.1580}          & 0.1521         & {\bf0.1470}         & 0.0392         & {\bf0.0389}         & 0.0256         & {\bf0.0253}         & 0.0297         &{\bf 0.0296}        \\
9                     & 0.3413         &{\bf 0.2858}         & 0.2304         & {\bf0.1948}         & 0.2517         & {\bf0.2090}         & 0.0222         & {\bf0.0214}         & 0.0189         & {\bf0.0183}         & 0.0116         &{\bf 0.0113}        \\
10                    & 0.0672         & {\bf0.0665}         & 0.0541         &{\bf 0.0537}         & 0.0397         & {\bf0.0390}         & 0.0207         & {\bf0.0200}           & 0.0152         & {\bf0.0151}         & 0.0140          &{\bf 0.0134}        \\
Avg.                  & 0.2931         &{\bf 0.2711}         & 0.2059         & {\bf0.1944}         & 0.2071         & {\bf0.1879}        & 0.0262         & {\bf0.0250}          & 0.0203         &{\bf 0.0198}         & 0.0159         & {\bf 0.0154}        \\ \hline
\end{tabular}
\end{table*}
}
 \begin{figure*}[!t]
\centering
\includegraphics[width=7in]{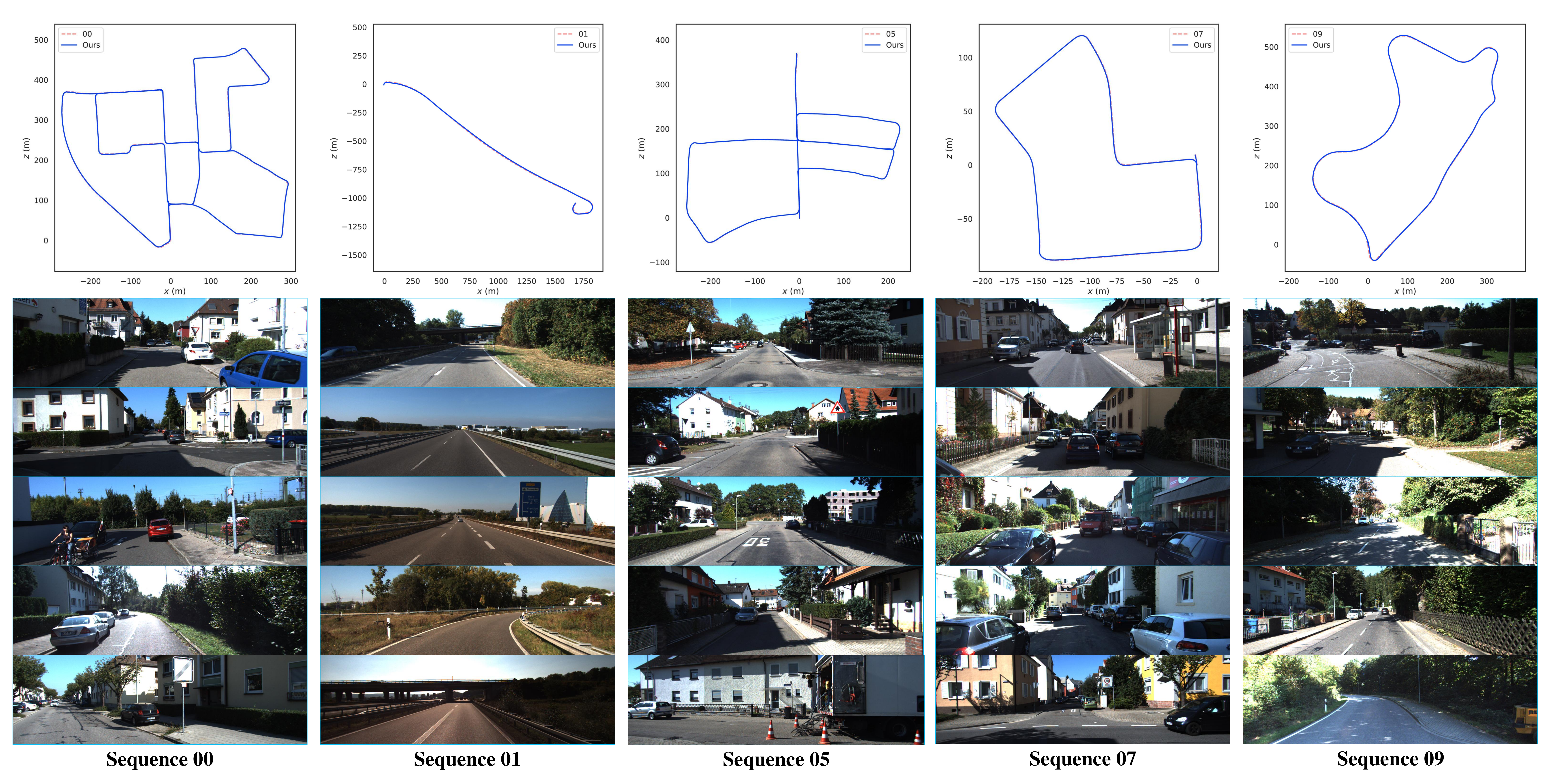}
\DeclareGraphicsExtensions.
\caption{Trajectory of our proposed algorithm results on sequences 00, 01, 05, 07, 09 of the KITTI dataset with five sample images for each sequence.}
\label{figure7}
\end{figure*}
  The results of our proposed method on KITTI dataset compared with ground-truth are given in Figure \ref{figure7} with 5 sample images of each sequence. It can be seen that our proposed algorithm achieved accurate results in various environments. Therefore, comparing with Salient-DSO only work in indoor and static environments, our proposed algorithm is more practical for applications. Even in indoor environment, our proposed algorithm also has super performance than most state of the art techniques (detailed in section \ref{section4.4}).

\subsection{Evaluation on EuRoc dataset}
\label{section4.4}
\begin{figure*}[htbp]
\centering
\includegraphics[width=7in]{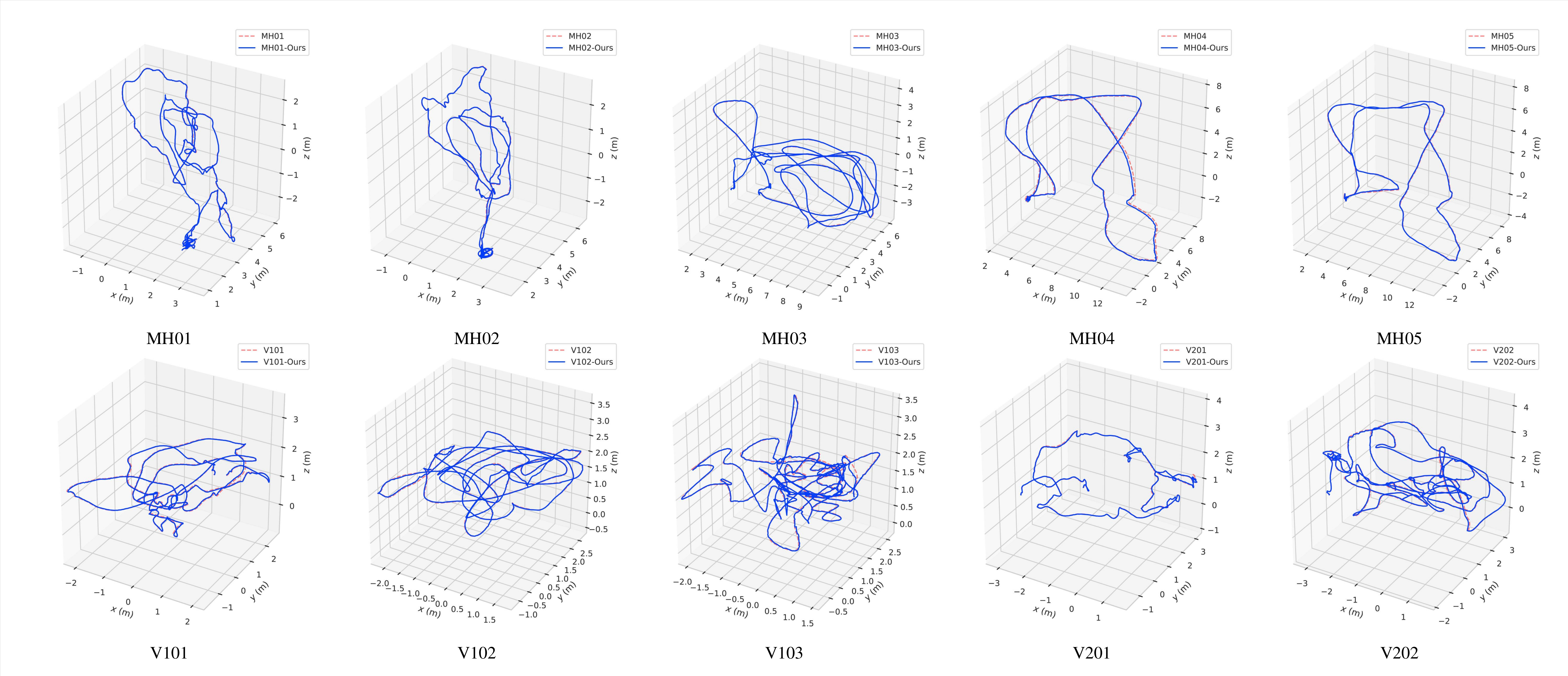}
\DeclareGraphicsExtensions.
\caption{Trajectory of our proposed algorithm results on sequences MH01-05, V101-03, and V201-02 of the RuRoc dataset.}
\label{figure8}
\end{figure*}

\begin{figure}[htbp]
\centering
\includegraphics[width=3.5in]{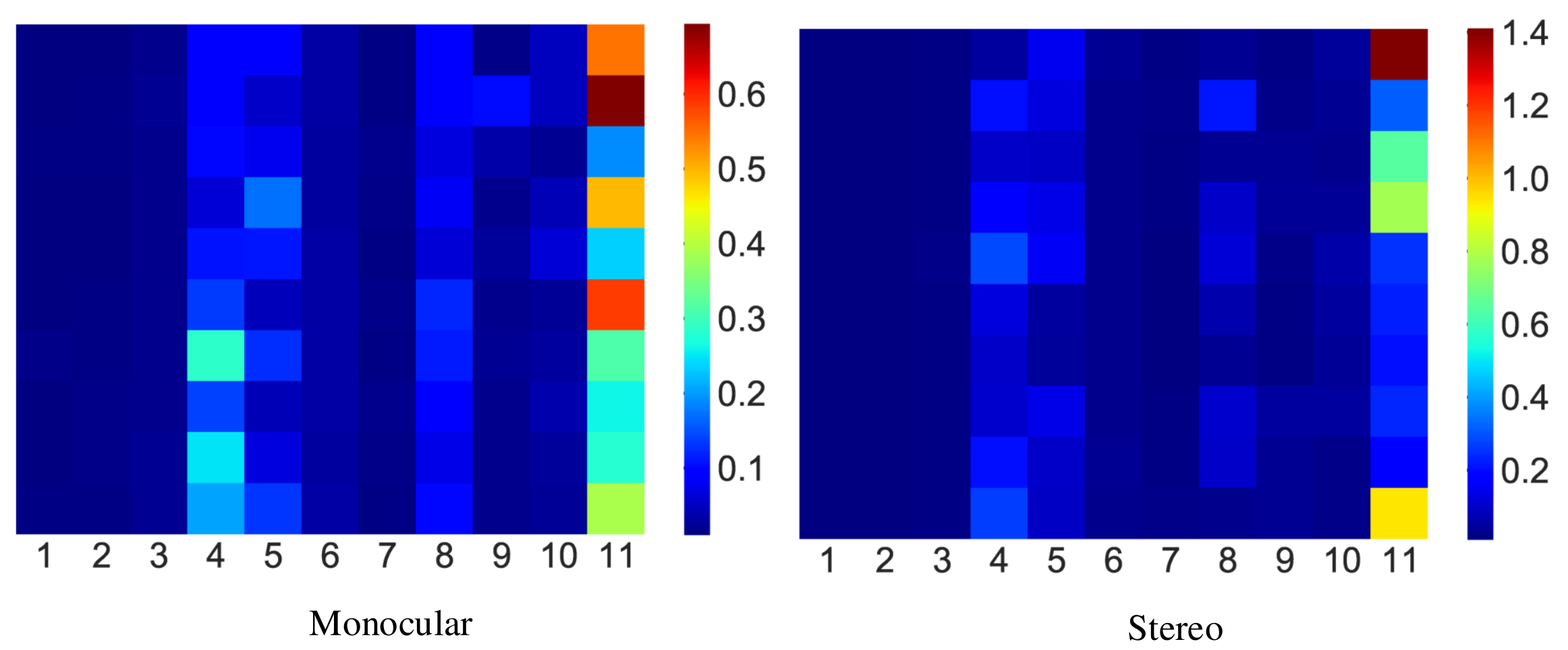}
\DeclareGraphicsExtensions.
\caption{Colored sequares represent the ATE for ten different execution in each sequence of the EuRoc dataset.}
\label{figure9}
\end{figure}

{\linespread{1.2}
\begin{table*}[]
\centering
\caption{The comparison of our proposed algorithm with other algorithms. }
\label{table3}
\begin{tabular}{cccccccccccc}
\hline
Sequences   & MH01   & MH02   & MH03   & MH04   & MH05   & V101   & V102   & V103   & V201   & V202   & V203   \\ \hline
Difficulty  & E      & E      & M      & D      & D      & E      & M      & D      & E      & M      & D      \\ \hline
\multicolumn{12}{c}{Monocular (ATE {[}m{]})}                                                                          \\ \hline
ORB-SLAM    & 0.0710  & 0.0670  & 0.0710  & 0.0820  & 0.0600   & {\bf 0.0150}  & 0.0200   & -      & 0.0210  & 0.0180  & -      \\
DSO         & 0.0460  & 0.0460  & 0.1720  & 3.8100   & 0.1100   & 0.0890  & 0.1070  & 0.9030  & 0.0440  & 0.1320  & 1.1520   \\
DSM         & 0.0390  & 0.0360  & 0.0550  & 0.0570  & 0.0670  & 0.0950  & 0.0590  & 0.0760  & 0.0560  & 0.0570  & 0.7840        \\
Salient-DSO & 0.0412 & 0.0435 & 0.1522 & 4.5629 & 0.0951 & 0.0963 & 0.6774 & 0.7562 & 0.0709 & 0.1322 & 1.2380        \\
ORB-SLAM3   & 0.0170  & {\bf 0.0170}  & 0.0310  & 0.0660  & 0.0440  & 0.0330  & 0.0160  & 0.0370  & 0.0210  & 0.0220  & -            \\
Ours        & {\bf 0.0144} & 0.0294 & {\bf 0.0261} & {\bf 0.0510}  & {\bf 0.0408} & 0.0328 &  {\bf 0.0124} & {\bf 0.0190}  & {\bf 0.0129} & {\bf 0.0165} & -           \\ \hline
\multicolumn{12}{c}{Stereo (ATE {[}m{]})}                                                                             \\ \hline
ORB-SLAM2   & 0.0350  & 0.0180  & 0.0280  & 0.1190  & 0.0600   & 0.0350  & 0.0200   & 0.0480  & 0.0370  & 0.0350  & --           \\
VINS-Fusion & 0.5400   & 0.4600   & 0.3300   & 0.7800   & 0.5000    & 0.5500   & 0.2300   & --     & 0.2300   & 0.2000    & --           \\
ORB-SLAM3   & 0.0250  & 0.0220  & 0.0270  & {\bf 0.0890}  & {\bf 0.0580}  & 0.0350  & 0.0210  & {\bf 0.0490}  & 0.0320  & {\bf 0.0270}  & 0.3610        \\
Ours        & {\bf 0.0137} & {\bf 0.0152} & { \bf 0.0207} & 0.1386 & 0.0878 & {\bf 0.0329} & {\bf 0.0169} & 0.0914 & {\bf 0.0213} & 0.0347 & {\bf 0.3526}       \\ \hline
\end{tabular}
\end{table*}
}
 In section, we evaluate our algorithm in indoor scenes. Different from section \ref{section4.3}, we measure accuracy with RMS ATE\cite{RN1050}, aligning the estimated trajectory with ground truth using a Sim(3) transformation in the pure monocular case. Moreover, we also compare our algorithm with other state of the art algorithms, such as ORB-SLAM\cite{RN632}, DSM\cite{RN1051}, DSO\cite{RN637}, salient-DSO\cite{RN781}, and ORB-SLAM3\cite{RN953}. Especially for Salient-DSO, this algorithm also uses saliency information to estimate pose. Since Salient-DSO cannot work in outdoor environments, we compare its performance in indoor environments to demonstrate the effectiveness of our proposed approach. The results are listed in Table \ref{table3}, and the correspond heatmaps are shown in Figure \ref{figure9} (each rectangle shows a different run). Some trajectory results on the EuRoc dataset with ground-truth are shown in Figure \ref{figure8}. 

 As shown in Table \ref{table3}, our proposed algorithm achieves more accurate results than state of the art techniques in literature in most sequences. We can also find that Salient-DSO is more accurate than DSO and ours is more accurate than ORB-SLAM3. Therefore, it convinces that combining saliency model with traditional monocular visual SLAM or VO can improve the performance. However, the saliency model used in Salient-DSO having center bias, this algorithm will track lost in outdoor environments due to there are no local salient objects for saliency prediction to work well. Contrast, our algorithm can work well in both indoor and outdoor environments, and achieve more accurate results in most sequences. However, our proposed algorithm performs poorly on difficult sequences due to fast motion and weak illumination. This is because our framework still uses the ORB-SLAM3 feature extraction method, and does not use the saliency information to improve the feature point extraction process, which is also a direction for our future work. Besides, the saliency prediction model used in this experiment is trained on Salient-KITTI dataset, which not contains indoor scenes. This may be another cause of poor results.

\section{Conclusion}
\label{section5}
In this article, we proposed a saliency-based SLAM framework, and the baseline comes from ORB-SLAM3. Inspired by the philosophy of saliency and attention, we use saliency prediction model to generate saliency information to mimic the human vision system. Based on the saliency information, we proposed an SBA method that can make the salient feature full play its value. We provide thorough qualitative evaluations on the KITTI dataset and EuRoc dataset to demonstrated that using salient information improves the accuracy and robustness. Besides, we also open-source our Salient-KITTI dataset. As a future research direction, we will propose a tightly coupled framework that uses saliency information to improve the feature extraction thread, even mapping thread. Besides, how to extract saliency information from IMU or Lidar data and used in SLAM/VO is also a future direction. 

\section*{Acknowledgment}
The authors thank the financial support of National Natural Science Foundation of China (Grant No: 51605054), Key Technical Innovation Projects of Chongqing Artificial Intelligent Technology (Grant No. cstc2017rgzn-zdyfX0039), Chongqing Social Science Planning Project (No:2018QNJJ16), Fundamental Research Funds for the Central Universities (No: 2019CDXYQC003).



\bibliographystyle{IEEEtran}


\bibliography{MyPaperReference}



%\vfill


\end{document}